\documentclass{article}
\usepackage{spconf,amsmath,amssymb,amsfonts,graphicx,multirow,cite}
\usepackage{hyperref}
\newcommand{\promptbox}[2]{%
  \par\smallskip\begingroup\setlength{\fboxsep}{2.5pt}\noindent\fbox{%
    \parbox{\dimexpr\linewidth-2\fboxsep-2\fboxrule\relax}{%
      \raggedright\textbf{System:} #1\par\vskip 0.5\smallskipamount
      \textbf{User:} #2}}\par\endgroup\smallskip}
\usepackage{xcolor}

\title{Transcribe, Translate, and Optimize: Joint Reward Learning for Speech Translation}

\name{Yanghe Dong$^{1}$ \qquad
      Wanting Huang$^{2}$ \qquad
      Weiran Wang$^{2}$}
\address{
$^{1}$Independent Researcher\\
\qquad \texttt{u7533843@alumni.anu.edu.au}\\
$^{2}$Department of Computer Science, University of Iowa, USA\\
\qquad \texttt{\{wanting-huang,weiran-wang\}@uiowa.edu}
}

\begin{document}
\ninept

\maketitle

\begin{abstract}
In LLM-based speech translation, transcription-based chain-of-thought (CoT) suffers from a mismatch between reference transcripts used in supervised fine-tuning (SFT) and model-generated transcripts at inference. To address this, we propose joint recognition and translation fine-tuning via group relative policy optimization (GRPO). We score both transcripts and translations, with translation conditioned on model-generated transcripts, and compare three token advantage strategies. Using Qwen2.5-Omni-3B across four languages, we evaluate CoT against direct speech translation (Direct ST) under SFT and GRPO, training on CoVoST~2 and testing on CoVoST~2 and FLEURS. CoT GRPO outperforms Direct ST GRPO by 1.77 and 0.83 average BLEU points on CoVoST~2 and FLEURS. Compared to CoT SFT, GRPO boosts BLEU by 0.82 and 0.67 points and reduces word error rate (WER) by 8.8\% and 7.2\% relatively. These results highlight reinforcement fine-tuning as an effective method to mitigate the training-inference mismatch, jointly improving recognition and translation.

\end{abstract}


\section{Introduction}

Speech translation (ST) converts source-language speech into target-language text. Cascaded systems combine automatic speech recognition (ASR) and machine translation, with early work using word lattices to pass multiple recognition hypotheses to translation~\cite{saleem_2004_lattice}. Early end-to-end models translate speech without generating intermediate transcripts at inference~\cite{berard_2016_listen,weiss_2017_seq2seq}. We call this output protocol \emph{direct speech translation} (Direct ST). ASR remains useful through transcription supervision and coupled decoders~\cite{weiss_2017_seq2seq,anastasopoulos_2018_tied}.

Some large language model (LLM)-based ST systems generate a transcript followed by a translation in one response~\cite{huang_2023_STLLM,luu_2025_E2EASRST}. We call this protocol \emph{transcription-based chain-of-thought speech translation} (CoT ST). Related work explores auxiliary ASR supervision~\cite{chen_2024_llast}, curriculum learning~\cite{du_2025_curriculum}, and phoneme-augmented CoT~\cite{gallego_2025_phoneme}. ASR hypotheses also guide translation through two-pass prompting~\cite{hu_2025_CoTST} or frozen-LLM representations supplied to a speech-conditioned decoder~\cite{higuchi_2025_E2ESTLLM}. Direct ST scales more consistently with pseudo-labeled data, while CoT achieves higher peak performance~\cite{pareras_2026_directSTLLM}.

For CoT ST, supervised fine-tuning (SFT) with reference transcript prefixes creates a training--inference mismatch, as translation follows potentially erroneous model-generated transcripts at inference. 
Speech-awareness studies demonstrate that CoT models exhibit a strong reliance on transcripts despite access to the audio signal~\cite{romerodiaz_2026_CoTST}. 
To address the prefix mismatch, prior work uses predicted~\cite{hu_2025_CoTST} or corrupted transcripts~\cite{romerodiaz_2026_CoTST} for supervised translation training.

Discriminative training can optimize task-level criteria over model-generated hypotheses. ASR systems have long used sequence-discriminative training after cross-entropy training~\cite{vesely_2013_sMBR}. Minimum word error rate (MWER) training more directly targets recognition errors by minimizing expected word errors over model hypotheses~\cite{prabhavalkar_2018_MWER,wang_2022_MWER}. Reinforcement learning (RL) similarly fine-tunes LLMs using rewards assigned to generated outputs.
In speech tasks, group relative policy optimization (GRPO)~\cite{shao_2024_grpo} has been applied to direct translation with BLEU rewards~\cite{elmakies_2026_grpoSU} and to recognition with WER, edit-distance, and exact-match rewards~\cite{shivakumar_2025_grpoASR}.~\cite{ren_2026_rlbr} applies GRPO to contextual ASR using rewards that prioritize biasing words. 

In this work, we investigate GRPO for CoT speech translation, optimizing translation conditioned on model-generated transcripts. Our main contributions are:
\begin{itemize}
    \setlength{\itemsep}{0pt}
    \setlength{\parsep}{0pt}
    \item We propose joint ASR--ST reward optimization to improve both recognition and translation within a single response generated by an audio large language model.

    \item We score both recognition and translation outputs during rollouts and investigate the Fully Coupled, Asymmetric, and Decoupled strategies for assigning token advantages to transcript and translation segments.

    \item Using Qwen2.5-Omni-3B and the same CoVoST~2 training examples, we empirically compare Direct and CoT ST under SFT and GRPO across four target languages. Evaluation on CoVoST~2 and FLEURS shows that CoT GRPO achieves the highest average BLEU and chrF++ among the compared systems and lowers WER relative to CoT SFT on both test sets.
\end{itemize}

\section{Background on LLM fine-tuning}
Let $x$ denote an input to an LLM, with desired response $o^*$.
SFT minimizes next-token cross-entropy over a reference response, effectively maximizing
$\log P(o^*|x)$.

GRPO~\cite{shao_2024_grpo} instead treats the LLM as a policy model $\pi_{\theta}$ and maximizes the expected reward of samples from the policy.
To calculate the objective, it first samples $G$ policy responses $\{o_i\}_{i=1}^{G}$ from an old policy $\pi_{\theta_{\mathrm{old}}}$, with rewards $\mathcal{R}=\{R_1,\ldots,R_G\}$. 
Then, the group-relative advantage for each response is computed as
\begin{equation}
\label{eq:group-advantage}
\hat{A}_i=\frac{R_i-\operatorname{mean}(\mathcal{R})}
{\operatorname{std}(\mathcal{R})}.
\end{equation}
For groups with zero reward variance, we set the advantages to zero. For token position $k$, we compute the policy probability ratio
\begin{equation}
\rho_{i,k}(\theta)=
\frac{\pi_\theta(o_{i,k}\mid x,o_{i,<k})}
{\pi_{\theta_{\mathrm{old}}}(o_{i,k}\mid x,o_{i,<k})}.
\end{equation}
For sequence-level rewards, GRPO assigns $\hat{A}_i$ to every token of response $i$. 
We utilize the DAPO variant of the GRPO implementation~\cite{yu_2025_dapo}, which incorporates asymmetric clipping and omits KL regularization.
For a single input and associated response group, the objective is to maximize
\begin{multline}
\label{eq:dapo-objective}
\mathcal{J}_{\mathrm{DAPO}}(\theta)
=\frac{1}{\sum_{i=1}^{G}|o_i|}
\sum_{i=1}^{G}\sum_{k=1}^{|o_i|}
\min\Bigl\{\rho_{i,k}\hat{A}_{i,k},\\
\operatorname{clip}(\rho_{i,k},1-\epsilon_{\mathrm{low}},
1+\epsilon_{\mathrm{high}})\hat{A}_{i,k}\Bigr\},
\end{multline}
where $\hat{A}_{i,k}$ is the advantage assigned to token $k$ of response $i$, $|o_i|$ is the response length in tokens and $\epsilon_{\mathrm{low}}$ and $\epsilon_{\mathrm{high}}$ set the clipping bounds for improved training stability. 

\section{Method}
\label{sec:method}

The input $x$ consists of a task prompt and an English speech utterance. We compare two protocols for RL fine-tuning: Direct ST (Sec~\ref{sec:method-direct}) outputs only the translation, whereas CoT ST (Sec~\ref{sec:method-transcript-first}) outputs the English transcript followed by the translation.

\subsection{Direct Speech Translation}
\label{sec:method-direct}

For Direct ST, we use the following prompt, where \texttt{\{target\}} is the target language.
\promptbox{You are a speech translation model.}{Listen to the provided English speech and produce a translation in \texttt{\{target\}} text.}
\noindent Let $y^*$ be the ground truth translation.
For GRPO, the raw reward for a sampled translation $y_i$ is
\begin{equation}
\label{eq:st-reward}
r_i^{\mathrm{ST}}=\frac{\mathcal{M}_{\mathrm{ST}}(y_i,y^*)}{100},
\qquad
R_i^{\mathrm{Direct}}=r_i^{\mathrm{ST}}.
\end{equation}
Here, $\mathcal{M}_{\mathrm{ST}}$ is sentence-level BLEU~\cite{papineni_2002_bleu} or chrF++~\cite{popovic_2017_chrfpp}, whose scores are scaled to $r_i^{\mathrm{ST}}\in [0,1]$. BLEU measures exact token $n$-gram precision with clipped match counts and a brevity penalty for overly short translations. chrF++ combines character $n$-grams with word unigrams and bigrams in a recall-weighted F-score. We compare the two reward metrics on the validation sets and retain chrF++, as reported in Section~\ref{sec:ablations}.

\subsection{Transcription-based CoT Speech Translation}
\label{sec:method-transcript-first}

For CoT ST, we use the following prompt. This schema is shared by SFT, GRPO, and inference.
\promptbox{You are a speech recognition and speech translation model.}{Listen to the provided English speech. First transcribe the speech in English, then translate it into \texttt{\{target\}}. Put the transcription inside \texttt{<transcription>}\allowbreak...\allowbreak\texttt{</transcription>}. Put the translation inside \texttt{<translation>}\allowbreak...\allowbreak\texttt{</translation>}.}

For SFT, the response contains ground truth $t^*$ and $y^*$ enclosed in their respective tags, and translation is trained with the reference transcript prefix. During GRPO rollouts and inference, translation instead follows the model-generated transcript within the same response. For GRPO, we extract the transcript $t_i$ and translation $y_i$ from sampled response $i$. The translation reward is computed based on $y_i$ following~\eqref{eq:st-reward}, and the ASR reward is
\begin{equation}
r_i^{\mathrm{ASR}}=\operatorname{clip}\!\left(1-\operatorname{WER}
(\mathcal{N}(t_i),\mathcal{N}(t^*)),0,1\right),
\end{equation}
where $\mathcal{N}$ is an English text normalizer (we use the Whisper normalizer~\cite{radford_2023_whisper} in this work). A binary format gate $g_i^{\mathrm{fmt}}$ equals $1$ only if the response contains exactly one nonempty transcript 
followed by one nonempty translation, each enclosed in the prescribed tags.

\par\vspace{0.05in}
\noindent\textbf{Token Advantage Assignment.}
We first construct gated joint (J), ASR, and ST rewards:
\begin{equation}
\begin{aligned}
R_i^{\mathrm{J}}&=g_i^{\mathrm{fmt}}(r_i^{\mathrm{ST}}+\lambda r_i^{\mathrm{ASR}}),\\
R_i^{\mathrm{ASR}}&=g_i^{\mathrm{fmt}} r_i^{\mathrm{ASR}},\qquad
R_i^{\mathrm{ST}}=g_i^{\mathrm{fmt}}r_i^{\mathrm{ST}},
\end{aligned}
\end{equation}
where $\lambda$ controls the ASR contribution to the joint reward. Each reward is normalized within its response group using~\eqref{eq:group-advantage}, yielding $\hat{A}_i^{\mathrm{J}}$, $\hat{A}_i^{\mathrm{ASR}}$, and $\hat{A}_i^{\mathrm{ST}}$, respectively. We compare three assignments:
\begin{itemize}
    \item \textbf{Fully Coupled}: All response tokens receive $\hat{A}_i^{\mathrm{J}}$.
    \item \textbf{Asymmetric}: Transcript tokens receive $\hat{A}_i^{\mathrm{J}}$, and translation tokens receive $\hat{A}_i^{\mathrm{ST}}$.
    \item \textbf{Decoupled}: Transcript tokens receive $\hat{A}_i^{\mathrm{ASR}}$, and translation tokens receive $\hat{A}_i^{\mathrm{ST}}$.
\end{itemize}
These token advantages are plugged into~\eqref{eq:dapo-objective} for training.

\subsection{Reference-Aware Training}
We evaluate reference-aware training for both protocols, following prior extensions of GRPO that incorporate reference responses~\cite{elmakies_2026_grpoSU,ren_2026_rlbr}. We append one reference response to each group of $G$ policy responses, using $y^*$ for Direct ST and the tagged $(t^*,y^*)$ response for CoT ST. All $G+1$ responses participate in reward normalization and the clipped objective.

\begin{table*}[!t]
\caption{Validation-set ablations with Fully Coupled advantages. In the Reference-aware block, 4 and 5 denote policy responses only, and 4 + 1 ref. includes one ground-truth response. Bold and underlined values indicate the best and second-best scores within each CoT block.}
\label{tab:validation-ablations}
\centering
\begin{tabular}{lll@{\hspace{12pt}}ccc@{\hspace{14pt}}ccc}
\hline
& & & \multicolumn{3}{c}{\textbf{CoVoST~2}} & \multicolumn{3}{c}{\textbf{FLEURS}} \\
\cline{4-6}\cline{7-9}
\textbf{Ablation} & \textbf{Protocol} & \textbf{Setting} & \textbf{WER} $\downarrow$ & \textbf{BLEU} $\uparrow$ & \textbf{chrF++} $\uparrow$ & \textbf{WER} $\downarrow$ & \textbf{BLEU} $\uparrow$ & \textbf{chrF++} $\uparrow$ \\
\hline
Base & Direct & --- & 5.74 & 31.65 & 40.40 & 7.40 & 25.77 & 36.53 \\
\hline
\multirow{2}{10em}{ST reward\\($\lambda=0.5$, 4 + 1 ref.)}
 & \multirow{2}{*}{CoT} & BLEU   & 5.35 & 34.64 & 42.89 & 5.44 & \textbf{28.53} & \textbf{39.19} \\
 &                     & chrF++ & \textbf{5.32} & \textbf{34.72} & \textbf{43.04} & \textbf{5.36} & 28.32 & 39.07 \\
\hline
\multirow{5}{10em}{Reference-aware\\($\lambda=0.5$, chrF++)}
 & \multirow{2}{*}{Direct} & 4          & 6.65 & 32.71 & 42.75 & 6.90 & 27.62 & 38.71 \\
 &                          & 4 + 1 ref. & 5.86 & 33.05 & 42.07 & 6.36 & 27.70 & 38.06 \\
\cline{2-9}
 & \multirow{3}{*}{CoT} & 4          & \underline{5.47} & 33.45 & \underline{44.05} & \underline{5.53} & \textbf{29.59} & \textbf{40.23} \\
 &                      & 5          & 5.60 & \underline{33.60} & \textbf{44.41} & 6.03 & 27.78 & \underline{39.60} \\
 &                      & 4 + 1 ref. & \textbf{5.32} & \textbf{34.72} & 43.04 & \textbf{5.36} & \underline{28.32} & 39.07 \\
\hline
\multirow{5}{10em}{ASR weight\\($G=4$, chrF++)}
 & \multirow{5}{*}{CoT} & 0.00 & 6.76 & \underline{34.02} & \textbf{44.58} & 6.76 & 28.83 & 39.98 \\
 &                      & 0.25 & 5.64 & 33.47 & \underline{44.13} & 5.63 & 28.57 & 40.00 \\
 &                      & 0.50 & \underline{5.47} & 33.45 & 44.05 & \underline{5.53} & \textbf{29.59} & \underline{40.23} \\
 &                      & 0.75 & \textbf{5.45} & \textbf{34.15} & 43.70 & \textbf{5.42} & 28.76 & \textbf{40.56} \\
 &                      & 1.00 & 5.48 & 33.29 & 43.93 & 5.60 & \underline{28.96} & 39.79 \\
\hline
\end{tabular}
\end{table*}

\section{Experiments}
\subsection{Data and Model}
All four fine-tuned configurations (SFT and GRPO for Direct and CoT translation) are initialized from the same base model: Qwen2.5-Omni-3B~\cite{xu_2025_qwen25omni}. Training is performed on CoVoST~2~\cite{wang_2021_covost2}. We assign each English recording to one of four target languages (Arabic, Chinese, German, and Japanese), avoiding repeated source recordings across target languages during training. The balanced training set contains 289,412 examples, totaling 430.12 hours of English speech, with 72,353 examples per target language.

We evaluate on all four target languages in the official CoVoST~2 validation and test sets, with 15,531 and 15,530 English recordings per target language (26.13 and 24.67 hours), respectively. For cross-corpus evaluation, we use FLEURS~\cite{conneau_2023_fleurs}, whose sentence IDs link parallel texts across languages. Within each split, we pair English recordings with target-language transcripts sharing the same sentence ID. This yields 1,509 validation and 2,403 test pairs from 394 and 647 unique English recordings (1.05 and 1.77 hours), respectively. FLEURS is not used for training.

\subsection{Training Configuration}
We perform full-parameter tuning with the ms-swift framework~\cite{zhao_2025_swift} on eight NVIDIA RTX PRO 6000 GPUs using AdamW~\cite{loshchilov_2019_adamw}. SFT uses a global batch size of 128, an initial learning rate of $10^{-5}$, cosine decay, and 5\% warmup. GRPO uses a constant learning rate of $10^{-6}$, and DAPO clipping parameters $\epsilon_{\mathrm{low}}=0.2$ and $\epsilon_{\mathrm{high}}=0.28$. Unless otherwise stated, we use $G=4$ without a reference. Each rollout samples responses for 32 prompts (128 policy responses; 160 for the $G=5$ ablation), at temperature 1.0 and top-$p$ 1.0. We perform one optimizer update using all sampled responses and any added references before collecting the next rollout. We train all models for one pass over the training set.

\subsection{Evaluation}
We use greedy decoding with temperature 0. Ablations and the separate comparison of advantage assignment settings use the CoVoST~2 and FLEURS validation sets.

Translation is evaluated under each system's native prompt using SacreBLEU~\cite{post_2018_sacrebleu} corpus BLEU and chrF++, reported per target language and as an unweighted mean. English WER is computed after Whisper English text normalization. For the base model and Direct ST systems, we run a separate ASR-only inference pass using the prompts below to obtain transcripts for WER evaluation. 
\promptbox{You are a speech recognition model.}{Transcribe the English audio into text without any punctuation marks.}
We also evaluate a base-model cascade without additional fine-tuning. It first transcribes the speech and then translates the resulting transcript with a text-only prompt using the same base model. Unlike CoT ST, its translation pass does not receive the speech input.

\begin{table}[!t]
\caption{Comparison of advantage assignment settings for CoT ST on the CoVoST~2 and FLEURS validation sets under a shared training configuration. The Fully Coupled results are repeated from the ASR-weight ablation at $\lambda=0.75$. Bold values mark the best score for each metric within each dataset, including ties.}
\label{tab:advantage-comparison}
\centering
\setlength{\tabcolsep}{3pt}
\begin{tabular}{@{}llccc@{}}
\hline
\textbf{Dataset} & \textbf{Setting} & \textbf{WER} $\downarrow$ & \textbf{BLEU} $\uparrow$ & \textbf{chrF++} $\uparrow$ \\
\hline
\multirow{3}{*}{CoVoST~2} & Fully Coupled & 5.45 & \textbf{34.15} & 43.70 \\
 & Asymmetric & 5.47 & 33.70 & \textbf{44.29} \\
 & Decoupled & \textbf{5.35} & 32.32 & 43.86 \\
\hline
\multirow{3}{*}{FLEURS} & Fully Coupled & \textbf{5.42} & \textbf{28.76} & \textbf{40.56} \\
 & Asymmetric & 5.49 & 28.62 & 39.97 \\
 & Decoupled & \textbf{5.42} & 26.88 & 39.39 \\
\hline
\end{tabular}
\end{table}

\begin{table*}[!t]
\caption{Test-set results on CoVoST~2 and FLEURS. Translation scores are reported for Arabic (ar), Chinese (zh), German (de), and Japanese (ja), and their unweighted average (Avg.). WER measures English transcription. Bold and underlined values indicate the best and second-best scores in each column within each dataset across all systems.}
\label{tab:main-test}
\centering
\setlength{\tabcolsep}{4pt}
\begin{tabular}{llc@{\hspace{14pt}}ccccc@{\hspace{14pt}}ccccc}
\hline
\multirow{2}{*}{\textbf{Model}} & \multirow{2}{*}{\textbf{Protocol}} & \multirow{2}{*}{\textbf{WER} $\downarrow$} & \multicolumn{5}{c}{\textbf{BLEU} $\uparrow$} & \multicolumn{5}{c}{\textbf{chrF++} $\uparrow$} \\
\cline{4-8}\cline{9-13}
 &  &  & ar & zh & de & ja & \textbf{Avg.} & ar & zh & de & ja & \textbf{Avg.} \\
\hline
\multicolumn{13}{l}{\textbf{CoVoST~2}} \\
\hline
\multirow{2}{*}{Base} & Direct & \multirow{2}{*}{7.71} & 18.70 & 41.14 & 28.35 & 29.77 & 29.49 & 43.24 & 27.67 & 52.68 & 26.87 & 37.61 \\
 & Cascade &  & 15.26 & 43.53 & 27.36 & 27.70 & 28.46 & 40.75 & 29.06 & 52.47 & 25.38 & 36.91 \\
\hline
\multirow{2}{*}{SFT} & Direct & 7.92 & 19.57 & 44.08 & 29.78 & 30.41 & 30.96 & 44.34 & 30.00 & 53.70 & 27.53 & 38.89 \\
 & CoT & \underline{7.17} & \underline{20.89} & \underline{46.42} & \underline{31.03} & \underline{32.05} & \underline{32.60} & 45.79 & 31.95 & \underline{54.95} & 28.69 & 40.34 \\
\hline
\multirow{2}{*}{GRPO} & Direct & 8.71 & 19.22 & \underline{46.42} & 29.19 & 31.77 & 31.65 & \underline{45.96} & \underline{31.97} & 54.51 & \underline{29.13} & \underline{40.39} \\
 & CoT & \textbf{6.54} & \textbf{21.72} & \textbf{47.74} & \textbf{31.55} & \textbf{32.67} & \textbf{33.42} & \textbf{47.84} & \textbf{32.66} & \textbf{56.51} & \textbf{30.74} & \textbf{41.94} \\
\hline
\multicolumn{13}{l}{\textbf{FLEURS}} \\
\hline
\multirow{2}{*}{Base} & Direct & \multirow{2}{*}{6.87} & 16.99 & 34.52 & 27.08 & 26.95 & 26.38 & 44.55 & 22.62 & 54.62 & 25.28 & 36.77 \\
 & Cascade &  & 14.27 & 36.94 & 24.22 & 25.18 & 25.15 & 41.66 & 23.95 & 52.82 & 24.24 & 35.67 \\
\hline
\multirow{2}{*}{SFT} & Direct & 6.39 & 17.81 & 38.92 & 27.60 & 27.57 & 27.98 & 45.52 & 26.90 & 55.10 & 26.04 & 38.39 \\
 & CoT & \underline{5.68} & \textbf{18.54} & 39.92 & 28.03 & \textbf{28.00} & \underline{28.62} & 46.37 & \underline{27.48} & 55.77 & 26.31 & 38.99 \\
\hline
\multirow{2}{*}{GRPO} & Direct & 8.13 & 17.36 & \underline{40.86} & \underline{28.04} & 27.57 & 28.46 & \underline{46.96} & 26.46 & \underline{56.90} & \underline{26.66} & \underline{39.25} \\
 & CoT & \textbf{5.27} & \underline{17.91} & \textbf{42.26} & \textbf{29.29} & \underline{27.70} & \textbf{29.29} & \textbf{47.37} & \textbf{27.86} & \textbf{57.60} & \textbf{27.60} & \textbf{40.11} \\
\hline
\end{tabular}
\end{table*}

\section{Results}

\subsection{Ablation Studies}
\label{sec:ablations}

Table~\ref{tab:validation-ablations} reports sequential validation ablations for our method. All ablation configurations improve BLEU and chrF++ over Base on both sets. All ablations use Fully Coupled advantages. Starting from $\lambda=0.5$ and $G=4$ with a reference, we compare the ST reward metric, reference-aware training and group size, and ASR reward weight, retaining the selections described below.

\subsubsection{Translation-Reward Metric Ablation}

The ST reward block of Table~\ref{tab:validation-ablations} compares BLEU and chrF++ rewards. Neither reward consistently outperforms the other across corpora. The chrF++ reward gives slightly higher translation scores on CoVoST~2, whereas BLEU gives slightly higher scores on FLEURS. We retain chrF++ given its lower observed WER on both corpora. %

\subsubsection{Reference-Aware Training Ablation}
With chrF++ reward and $\lambda=0.5$, we retain $G=4$ without a reference for subsequent ablations (Table~\ref{tab:validation-ablations}, Reference-aware). This setting has higher FLEURS BLEU and chrF++ and lower WER on both corpora than $G=5$ without a reference, although its CoVoST~2 translation scores are slightly lower. Relative to $G=4$ with an appended reference, it has higher CoVoST~2 chrF++ and higher FLEURS BLEU and chrF++, each by over one point, but lower CoVoST~2 BLEU and slightly higher WER on both corpora.

For Direct ST, reference-aware training lowers standalone ASR WER on both corpora, while its effects on translation are mixed.

\subsubsection{ASR-Reward Weight Ablation}
Table~\ref{tab:validation-ablations} (ASR weight) reports this ablation. Adding an ASR reward substantially reduces WER, with most of the reduction already present at $\lambda=0.25$.
This supports using an explicit ASR reward to improve transcription accuracy during joint ASR--ST training.

The effects of the ASR reward on translation are mixed across metrics and corpora. Within this ablation, we retain $\lambda=0.75$, which gives the lowest observed WER on both corpora alongside the highest CoVoST~2 BLEU and FLEURS chrF++.

\subsection{Comparison of Advantage Assignment Settings}
\label{sec:advantage-comparison}
Table~\ref{tab:advantage-comparison} compares token advantage assignment strategies for CoT ST on the validation sets. All three use chrF++ ST reward and $G=4$. Fully Coupled and Asymmetric use $\lambda=0.75$. These settings are taken from the Fully Coupled ablations without additional tuning.

Under the shared configuration, Fully Coupled achieves the highest BLEU on both validation sets and the highest chrF++ on FLEURS. Compared with Fully Coupled, Asymmetric improves CoVoST~2 chrF++, while Decoupled lowers CoVoST~2 WER, both at the cost of lower BLEU on both corpora.

\subsection{Main Results on Test Sets}

We use the Fully Coupled setting evaluated in Section~\ref{sec:advantage-comparison} for CoT GRPO test evaluation. The validation ablations in Section~\ref{sec:ablations} determine the remaining configuration: chrF++ reward, $\lambda=0.75$, and $G=4$ without a reference.

Table~\ref{tab:main-test} reports the test results. CoT GRPO achieves the highest average BLEU and chrF++ on both corpora and leads all per-language translation metrics except FLEURS Arabic and Japanese BLEU, where it ranks second. Its average BLEU also exceeds Direct ST GRPO by 1.77 points on CoVoST~2 and 0.83 on FLEURS.%

Relative to CoT SFT, GRPO improves average BLEU by 0.82 and 0.67 on CoVoST~2 and FLEURS, respectively, while also increasing chrF++ on both corpora. Relative WER reductions are 8.8\% (from 7.17\% to 6.54\%) on CoVoST~2 and 7.2\% (from 5.68\% to 5.27\%) on FLEURS. In contrast, Direct ST GRPO improves average translation scores over Direct ST SFT but worsens WER. These results support joint ASR--ST optimization for improving both translation quality and recognition accuracy.
For the base model, Direct ST achieves higher average BLEU and chrF++ than the cascade on both test sets. Only Chinese favors the cascade in both metrics.

\begin{table}[t]
\caption{Transcript intervention on test sets, averaged over four target languages. Self-generated WERs (SFT/GRPO) are 7.17\%/6.54\% on CoVoST~2 and 5.68\%/5.27\% on FLEURS.}
\label{tab:transcript-intervention}
\centering
\setlength{\tabcolsep}{2pt}
\begin{tabular}{@{}llcccc@{}}
\hline
 & & \multicolumn{2}{c}{\textbf{CoVoST~2}} & \multicolumn{2}{c}{\textbf{FLEURS}} \\
\cline{3-4}\cline{5-6}
\textbf{Model} & \textbf{Transcript} & \textbf{BLEU} $\uparrow$ & \textbf{chrF++} $\uparrow$ & \textbf{BLEU} $\uparrow$ & \textbf{chrF++} $\uparrow$ \\
\hline
\multirow{2}{*}{SFT} & Self-generated & 32.60 & 40.34 & 28.62 & 38.99 \\
 & Ground truth & 35.15 & 43.12 & 31.21 & 40.74 \\
\hline
\multirow{2}{*}{GRPO} & Self-generated & 33.42 & 41.94 & 29.29 & 40.11 \\
 & Ground truth & 36.19 & 44.90 & 31.99 & 42.12 \\
\hline
\end{tabular}
\end{table}

\subsection{Transcript Intervention Analysis}

We test whether the translation advantage of CoT GRPO persists with identical reference transcripts. Using the SFT and GRPO test checkpoints, we retain the speech input and CoT prompt but provide the reference transcript span and opening translation tag as an assistant prefix, then generate the translation using greedy decoding. The self-generated condition uses the native CoT results from Table~\ref{tab:main-test}.

Table~\ref{tab:transcript-intervention} shows that reference transcripts improve average BLEU and chrF++ on both corpora. GRPO shows slightly larger gains than SFT in both BLEU and chrF++. Thus, even after joint reward optimization, translation can still benefit from %
a more accurate transcript. GRPO retains a translation advantage under identical reference transcripts, suggesting it can better exploit the transcript for translation.

\section{Conclusion}

We studied GRPO for CoT speech translation with joint rewards for generated transcripts and translations. In both in-domain and cross-corpus evaluation, CoT GRPO improves average translation scores over Direct ST GRPO, while improving both average translation scores and recognition over CoT SFT. Validation ablations support an explicit ASR reward for transcription accuracy, and Fully Coupled advantages yield the highest BLEU on both corpora under the shared configuration. Future work will evaluate whether these findings extend to other model scales and language directions.

\section{Compliance with Ethical Standards}

This study uses the publicly available CoVoST~2 and FLEURS
datasets for experiments.

\section{Acknowledgments}

The authors have no relevant financial or nonfinancial interests to disclose.

\bibliographystyle{IEEEbib}
\bibliography{references}

\end{document}